\documentclass[letterpaper, 10 pt, conference]{ieeeconf}  % Comment this line out if you need a4paper

\IEEEoverridecommandlockouts                              % This command is only needed if
\usepackage{amsmath,amssymb}
\usepackage{booktabs}
\usepackage{graphicx}
\usepackage[table]{xcolor}
\usepackage{multirow}

\title{\LARGE \bf
DRIFT: Drift and Aggregation for Motion Planning
}

\author{Yining Xing$^{1}$, Zhiyuan Liu$^{1}$, Zehong Ke$^{1}$, Wenhao Yu$^{2,*}$, and Jianqiang Wang$^{1}$%
\thanks{$^{1}$Yining Xing, Zhiyuan Liu, Zehong Ke, and Jianqiang Wang are with the School of Vehicle and Mobility, Tsinghua University, Beijing, China.}%
\thanks{$^{2}$Wenhao Yu is with the State Key Laboratory of Intelligent Green Vehicle and Mobility, Tsinghua University, Beijing, China (Corresponding author, e-mail: eyre530056@gmail.com).}%
\thanks{This work was supported by National Key R\&D Program of China: 2023YFB2504400, National Natural Science Foundation of China grant 52572477, and Independent Research Project of the State Key Laboratory of Intelligent Green Vehicle and Mobility, Tsinghua University (No. ZZ-PY-20250409).}}

\begin{document}

\maketitle
\thispagestyle{empty}
\pagestyle{empty}

%%%%%%%%%%%%%%%%%%%%%%%%%%%%%%%%%%%%%%%%%%%%%%%%%%%%%%%%%%%%%%%%%%%%%%%%%%%%%%%%
\begin{abstract}

End-to-end trajectory planners need to represent multiple plausible driving behaviors while producing a single executable trajectory under real-time constraints. Proposal-based approaches address this ambiguity by generating multiple candidates, but converting the proposal set into a final plan remains a key design problem. We present DRIFT, a fixed-depth planner that combines one-step drifting in a compact trajectory latent space with scene-aware proposal aggregation. Conditioned on features from a pretrained visual encoder, the DRIFT Decoder generates 48 proposal features in a single batched pass, with 32 samples at $\alpha=0.5$ and 16 samples at $\alpha=0.9$. A lightweight Aggregation Head integrates these features with scene, navigation, and ego-state information and directly predicts the final trajectory without requiring trajectory-level quality labels for aggregation. Its output is trained with expert-trajectory imitation and a map-derived boundary regularizer that penalizes waypoints outside the drivable polygon and inside waypoints near its boundary. On NAVSIM navtest, DRIFT achieves 89.6 PDMS and 90.4 EPDMS, with strong drivable-area compliance and ego progress among the methods compared. The proposal-generation and aggregation module runs in 10.82 ms on an NVIDIA RTX 4090, while full-model inference including the visual backbone takes 66.43 ms. These results show that one-step latent proposal generation and direct aggregation provide an efficient design for multi-hypothesis motion planning.

\end{abstract}


%%%%%%%%%%%%%%%%%%%%%%%%%%%%%%%%%%%%%%%%%%%%%%%%%%%%%%%%%%%%%%%%%%%%%%%%%%%%%%%%
\section{INTRODUCTION}
\label{sec:intro}

End-to-end (E2E) autonomous driving systems~\cite{wu2022trajectory,hu2023planning} learn to predict ego motion directly from sensor observations and navigation context. Compared with traditional modular pipelines~\cite{dauner2023parting,chen2024end}, this formulation reduces reliance on hand-designed interfaces between perception and planning and allows scene representations to be optimized for the planning objective. Nevertheless, a unified model does not make future driving behavior deterministic. At intersections, during lane changes, or when interacting with other road users, the same observed scene may admit several plausible ego maneuvers. An effective planner should therefore preserve multiple motion hypotheses before committing to a single executable trajectory.

Representing multiple hypotheses introduces an additional computational challenge, because candidate generation must remain compatible with real-time planning. Anchor-based~\cite{madjid2025trajectory} and DETR-style set prediction methods~\cite{casas2024detra} provide efficient candidate generation, but fixed anchors or set queries can limit flexibility and long-horizon coverage. Generative planners based on diffusion~\cite{liao2025diffusiondrive,zheng2025diffusion} or autoregressive decoding~\cite{yang2024genad} offer flexible candidate generation but commonly rely on iterative inference. MeanFuser~\cite{wang2026meanfuser} recently demonstrated a one-step alternative based on MeanFlow and Gaussian mixture noise. DRIFT likewise avoids iterative inference, but performs generative drift in a compact PCA trajectory latent space rather than generating proposals directly in physical trajectory space. Its decoder produces multiple proposal features in a single batched pass.

Generating proposals alone does not determine what the vehicle should execute. A planner still has to combine the proposal set with the current scene, route command, and ego state, then output a single trajectory with stable behavior across frames. Learned trajectory scoring provides one solution, either by selecting the highest-scoring candidate or by using scores to guide iterative proposal evolution~\cite{li2025generalized,ke2026grade}. Another solution is direct reconstruction, as demonstrated by MeanFuser, which integrates its sampled physical trajectories with an Adaptive Reconstruction Module trained from expert demonstrations~\cite{wang2026meanfuser}. DRIFT follows this direct-reconstruction direction with a different proposal interface. It pools latent proposal features before PCA decoding and uses scene, navigation, and ego-state information to predict the final trajectory.

Scorer-based and direct-reconstruction designs also differ in how the proposal-to-plan step is supervised. Learned scorers can be trained with trajectory-level quality targets derived from planning metrics or other evaluation objectives~\cite{ke2026grade,wang2026drive}. Such targets provide explicit guidance for candidate selection or refinement. DRIFT instead trains the Aggregation Head with expert-trajectory imitation and map-derived boundary regularization. Neither objective requires trajectory-quality labels, and no separate scorer is used at inference time.

The final trajectory also needs geometric supervision in physical space. Imitation losses align the prediction with the logged expert, but they do not explicitly encode clearance from drivable-area boundaries~\cite{jia2023driveadapter,wang2025comdrive}. We therefore add a map-derived boundary regularizer. Waypoints outside the drivable polygon receive a fixed penalty, and inside waypoints are penalized according to their distance from the nearest boundary. The term is used as a local geometric regularizer, not as a formal safety guarantee.

We present DRIFT, an end-to-end planner built on a pretrained visual backbone~\cite{bardes2024revisiting}. On NAVSIM navtest~\cite{Dauner2024NEURIPS}, DRIFT achieves 89.6 PDMS and 90.4 EPDMS among the methods compared in this work. Its trajectory generation module keeps a fixed-depth computation profile, generates 48 intermediate proposals, aggregates them into one final trajectory, and runs in 10.82 ms on an RTX 4090. Our contributions are summarized below.
\begin{itemize}
    \item We adapt one-step drifting to a compact trajectory latent space and generate a fixed-size set of scene-conditioned proposals without iterative denoising.
    \item We develop a scene-aware Aggregation Head that pools latent proposals across drift intensities before physical decoding and is trained without trajectory-quality labels.
    \item We combine this design with a map-derived boundary regularizer and evaluate the integrated planner on NAVSIM navtest, obtaining 89.6 PDMS and 90.4 EPDMS.
\end{itemize}

%===============================================================================

\section{RELATED WORK}
\label{sec:related}

\subsection{End-to-End Autonomous Driving}

End-to-end autonomous driving encompasses several model organizations rather than a single monolithic design. Imitation-based planners learn driving outputs directly from sensor observations, ego state, and navigation context, with performance depending on the chosen representation, supervision, and evaluation protocol~\cite{dauner2023parting,chen2024end}. TransFuser uses Transformer-based camera and LiDAR fusion to combine geometric and visual evidence for imitation-based driving~\cite{chitta2022transfuser}. TCP instead predicts both a future trajectory and low-level control, allowing the two output branches to provide complementary planning signals~\cite{wu2022trajectory}. These approaches illustrate that end-to-end learning can retain different internal structures even when the overall system is optimized for driving.

Another line of work expands the scope of joint optimization. UniAD organizes BEV perception, tracking, motion forecasting, and planning within a planning-oriented query framework~\cite{hu2023planning}. Such unification can facilitate information sharing across tasks, but it also makes the interface between perception and planning an important architectural choice. DriveAdapter explicitly studies this coupling and adapts pretrained perception features for planning~\cite{jia2023driveadapter}, while DriveTransformer explores a shared Transformer formulation for scalable end-to-end driving~\cite{jiadrivetransformer}. Together, these systems span different choices in sensor fusion, task sharing, module coupling, and the allocation of trainable capacity.

Pretrained visual representations provide a complementary way to allocate model capacity. V-JEPA learns video features by predicting representations in latent space rather than reconstructing pixels~\cite{bardes2024revisiting}, and Drive-JEPA applies video-JEPA representations to end-to-end driving with multimodal trajectory distillation~\cite{wang2026drive}. DRIFT adopts a representation-first configuration in which a V-JEPA-pretrained ViT encodes the front-view image history while the trainable planning modules concentrate on trajectory proposal generation and scene-aware aggregation. Its focus is therefore not a new sensor-fusion backbone or a fully unified perception and prediction stack, but the planner built on top of pretrained scene features. The following subsection discusses the orthogonal problem of generating or refining multiple trajectory hypotheses and converting them into one executable plan.

\subsection{Trajectory Proposal Generation and Refinement}

Trajectory planners often need several candidates because a single regressed trajectory can average over ambiguous decisions~\cite{madjid2025trajectory}. Anchor-based methods~\cite{madjid2025trajectory} are efficient, but static anchors can limit behavior coverage outside the training distribution. DETR-style set prediction~\cite{casas2024detra} removes fixed priors through bipartite matching, yet long-horizon coverage remains difficult. Generative planners, including diffusion-based models~\cite{liao2025diffusiondrive,zheng2025diffusion} and autoregressive decoders~\cite{yang2024genad}, provide flexible candidate generation but usually require iterative inference. Closest to DRIFT, MeanFuser~\cite{wang2026meanfuser} uses one-step MeanFlow sampling from Gaussian mixture noise to generate physical trajectory proposals and an attention-based Adaptive Reconstruction Module to produce the final plan from expert supervision. DRIFT shares this one-step proposal-and-integration structure, while applying distributional drifting in a PCA latent space and aggregating proposal features before physical decoding.

Iterative refinement is another route to high-quality trajectories. Gameformer~\cite{huang2023gameformer} modeled trajectory prediction as a level-$k$ game in which agents update their plans based on other agents' anticipated behavior. DiffRefiner~\cite{yin2026diffrefiner} used diffusion-based denoising conditioned on scene features. HiPro-AD~\cite{chen2025hipro} used sparse Transformers with hybrid attention for proposal refinement. DRIFT avoids iterative refinement at inference time. It performs conditional drift in latent feature space, pools the resulting proposals through learned attention, and decodes a single trajectory with scene features.

%===============================================================================

\begin{figure*}[t]
  \centering
  \includegraphics[width=0.7\linewidth]{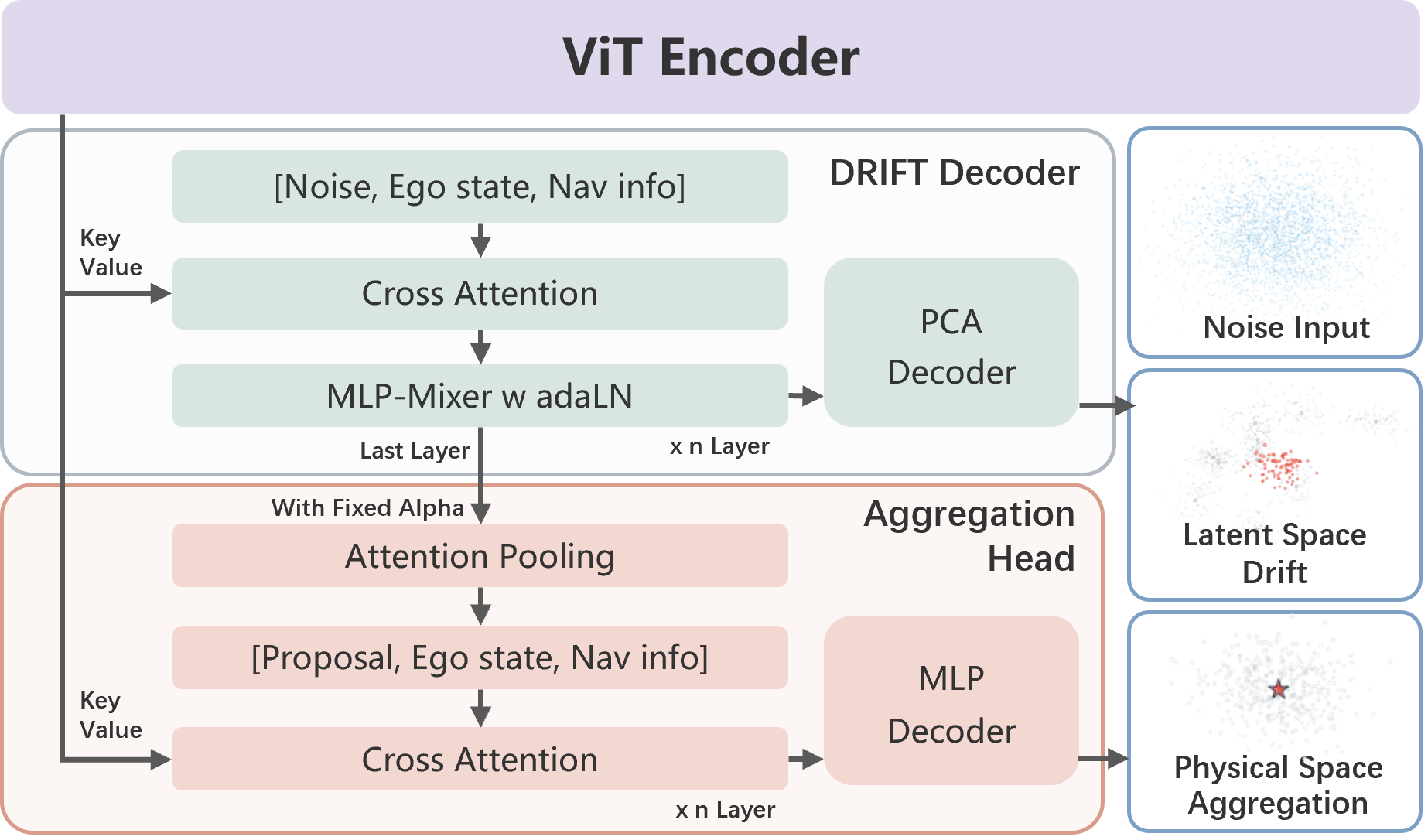}
  \caption{Overview of the DRIFT architecture. A V-JEPA pretrained ViT encodes front-view camera frames into shared scene features. The DRIFT Decoder performs conditional drift from isotropic noise to generate multi-hypothesis intent proposals. The Aggregation Head fuses these proposals with navigation and ego-state information through cross-attention and outputs a single executable trajectory.}
  \label{fig:framework}
\end{figure*}
\section{METHOD}
\label{sec:method}

Given historical and current front-view camera frames, high-level navigation commands, and the ego-state vector $\mathbf{s}_{\text{ego}}=(v_x,v_y,a_x,a_y)$, DRIFT predicts the ego vehicle's executable trajectory over the next 4 seconds. This section describes the architecture and the latent-space drift formulation.

\subsection{Overall Architecture}
\label{subsec:overall_architecture}

Fig.~\ref{fig:framework} shows the three-stage DRIFT pipeline. A ViT encoder pretrained with V-JEPA extracts scene features from current and past front-view camera frames. The DRIFT Decoder takes Gaussian noise and performs conditional drift in latent feature space by cross-attending to these scene features. It outputs hidden proposal features that represent different driving intents, while a learned MLP maps each feature to PCA coefficients for drift supervision and proposal visualization. The hidden proposal features are pooled by learned attention, concatenated with the navigation command and ego-vehicle state, and refined through cross-attention with the scene features. A final MLP head decodes the result into waypoints $\hat{\mathbf{T}}_{\text{final}} = \{(\hat{x}_t, \hat{y}_t, \hat{\theta}_t)\}_{t=1}^{T}$, where each waypoint contains position and heading.

\subsection{DRIFT Decoder}
\label{subsec:draft_decoder}

The DRIFT Decoder is a conditional trajectory generator. It starts from a Gaussian noise vector, treated as the trajectory token, and conditions generation on scene features and ego state. The trajectory token is concatenated with a navigation command token and an ego-state token encoding $\mathbf{s}_{\text{ego}}$ to form the decoder token sequence. A cascade of Decoder Blocks processes this sequence.

Each Decoder Block has two operations. The trajectory token first cross-attends to the ViT scene features, and the result is added back through a residual connection. The full decoder token sequence then passes through an MLP-Mixer. Token mixing and channel mixing allow the trajectory token to exchange information with the navigation and ego-state tokens. The drift intensity factor $\alpha$ is injected into the Mixer through Adaptive Layer Normalization (adaLN), which is defined as
\begin{equation}
\text{adaLN}(x, \alpha) = \gamma(\alpha) \odot \text{LayerNorm}(x) + \beta(\alpha)
\label{eq:adaln}
\end{equation}
where $\gamma(\alpha)$ and $\beta(\alpha)$ are affine parameters regressed from an $\alpha$-embedding. This modulation lets one decoder cover exploratory settings ($\alpha \to 0$) and imitation-biased settings ($\alpha \to 1$).

Directly applying the RBF-based drift to concatenated waypoint coordinates places the affinity computation in a relatively high-dimensional physical space. In our preliminary experiments, the resulting proposal distribution tended to fragment into isolated local clusters. We therefore use separate hidden-feature and PCA-coefficient representations. For the $k$-th noise sample, the Decoder produces a hidden proposal feature $\mathbf h_k$, which a learned MLP $f_\psi$ maps to PCA coefficients $\mathbf x_k$. The fixed inverse PCA transform then reconstructs the position-only proposal $\hat{\mathbf p}_k$ according to
\begin{equation}
\begin{aligned}
\mathbf h_k &= D_\phi(\boldsymbol\epsilon_k,\mathbf c,\alpha),
&\mathbf x_k &= f_\psi(\mathbf h_k),\\
\hat{\mathbf p}_k &= \boldsymbol\mu+\mathbf U\mathbf x_k,
\end{aligned}
\label{eq:proposal_path}
\end{equation}
where $\mathbf c$ contains the scene, navigation, and ego-state conditions, $\mathbf x_k\in\mathbb R^d$, $\hat{\mathbf p}_k\in\mathbb R^{2T}$ contains only $(x,y)$ waypoint coordinates, and $\mathbf U$ and $\boldsymbol\mu$ are the fixed PCA basis and mean. The PCA basis is fitted offline on real driving trajectories. Drift supervision is applied to $\mathbf x_k$, whereas the Aggregation Head consumes $\mathbf h_k$ directly.

\subsection{Latent Feature Space Drift}
\label{subsec:latent_drift}

The decoder is supervised with a drifting mechanism adapted from Drifting Models~\cite{deng2026generative}. Drifting Models describe generation as a distribution-level evolution in which generated samples are attracted toward real data and repelled from one another by a drifting field. At equilibrium, the field vanishes, which enables one-step generation. We adapt this mechanism to trajectory planning by applying the drift in the PCA latent space described above.

Using the notation in Eq.~\ref{eq:proposal_path}, $\mathbf x_k\in\mathbb R^d$ is the PCA-space representation of the $k$-th generated proposal. We compute its affinity to positive samples $\{y_m^+\}$ and negative samples $\{y_n^-\}$ with an RBF kernel at temperature $\tau$. The negative set consists of the other generated proposals. Sinkhorn symmetrization gives a doubly normalized affinity matrix $A$. The drift vector at temperature $\tau$ is the displacement from $\mathbf x_k$ toward the affinity-weighted positive centroid minus the displacement toward the negative centroid and is written as
\begin{equation}
\begin{aligned}
\mathbf{V}_\tau(\mathbf{x}_k)
&= \sum_{m} \frac{A_{k,m}^+}{\sum_{m} A_{k,m}^+}
   (\mathbf{y}_m^+ - \mathbf{x}_k) \\
&\quad - \sum_{n} \frac{A_{k,n}^-}{\sum_{n} A_{k,n}^-}
   (\mathbf{x}_n - \mathbf{x}_k)
\end{aligned}
\end{equation}
We sum this field over multiple temperatures $\{\tau_s\}$. Using the expert trajectory as the positive set gives the conditional drift $\mathbf{V}^{\text{cond}}$. Using trajectories from other scenes that satisfy the current road topology gives the unconditional drift $\mathbf{V}^{\text{unc}}$. A drift intensity factor $\alpha \in [0, 1]$ interpolates between them according to
\begin{equation}
\mathbf{V}_k^{\text{total}} = \mathbf{V}_k^{\text{unc}} + \alpha \cdot (\mathbf{V}_k^{\text{cond}} - \mathbf{V}_k^{\text{unc}})
\label{eq:latent_drift}
\end{equation}
When $\alpha \to 0$, the unconditional prior dominates and encourages exploration under the current road topology. When $\alpha \to 1$, the field is pulled toward the expert demonstration. Following Drifting Models~\cite{deng2026generative}, the training objective minimizes $\|\mathbf{x}_k - \text{sg}(\mathbf{x}_k + \mathbf{V}_k^{\text{total}})\|^2$, where $\text{sg}$ denotes stop-gradient. At inference time, different $\alpha$ values and noise samples produce a fixed-size set of exploratory and imitation-biased hypotheses.

\subsection{Scene-Aware Aggregation}
\label{subsec:aggregation}

The Scene-Aware Aggregation module fuses the proposal set from the DRIFT Decoder into a single executable trajectory.

At inference, the Decoder is queried with a fixed proposal set. We use two drift intensity factors, $\alpha=0.5$ and $\alpha=0.9$. The decoder samples 32 Gaussian noise vectors at $\alpha=0.5$ and 16 at $\alpha=0.9$, producing 48 hidden proposal features $\{\mathbf h_k\}$ in one batched pass. For each $\alpha$, a learnable query token cross-attends to the hidden features in that group and pools them into one representative token. The pooled tokens from both $\alpha$ values form a compact proposal sequence.

The proposal tokens are concatenated with the navigation command token and ego-state token to form a query sequence. A cascade of Aggregation Blocks (AggBlocks) processes this sequence. In each AggBlock, the query tokens attend to the encoded ViT scene features through gated cross-attention. A sigmoid gate on the attention output controls how much scene information enters the query tokens. The tokens then pass through an MLP-Mixer, where token and channel mixing exchange information across proposal, navigation, and ego-state tokens. An MLP compresses the final token sequence into a single query vector, with a residual connection from the mean input query.

After the final AggBlock, a LayerNorm and MLP head decode the query vector into $T$ future waypoints with position $(x, y)$ and heading $\theta$. The heading is predicted as $\hat{\theta}_t = \pi \cdot \tanh(z_t)$, where $z_t$ is the raw network output at timestep $t$. In NAVSIM's short-horizon ego-centric setting, heading changes are locally continuous in typical maneuvers, so the heading component is supervised together with waypoint positions in the imitation loss.

The final plan follows a separate decoding path. Rather than selecting one of the position proposals $\hat{\mathbf p}_k$ or aggregating their PCA coefficients $\mathbf x_k$, the Aggregation Head consumes the hidden proposal features $\mathbf h_k$ before the PCA-coefficient MLP $f_\psi$ and directly regresses $(x,y,\theta)$ with a learned MLP. This direct head predicts heading explicitly and allows the final plan to depart from the fixed PCA subspace, thereby preventing PCA truncation and reconstruction errors from propagating to the executable trajectory. PCA therefore parameterizes the proposal-generation branch without constraining the final output. The pooling weights and output head are optimized by the trajectory imitation and road-boundary objectives, neither of which requires trajectory-quality labels or evaluator scores.

\subsection{Physical-Space Safety Losses}
\label{subsec:physical_drift}

Latent-space drift shapes the proposal distribution. The final aggregated trajectory is supervised in physical coordinates to improve execution accuracy and drivable-area compliance. This physical-space supervision has two losses.

Following Drive-JEPA~\cite{wang2026drive}, we use an arc-length-weighted L1 loss. The loss reweights samples in a batch according to the length of their ground-truth trajectories. For each sample, the per-timestep L1 error is averaged across waypoints. The sample weight is $w = 1/(\gamma + \ell)$, where $\ell = \sum_t \|\mathbf{T}_{t+1}^{\text{GT}} - \mathbf{T}_t^{\text{GT}}\|$ is the ground-truth arc length and $\gamma$ is a smoothing constant. Short trajectories, such as low-speed or congested driving, receive higher weights. We normalize the weights to have mean one in each batch and define the imitation loss as
\begin{equation}
\mathcal{L}_{\text{traj}} = \frac{1}{B} \sum_{b=1}^{B} w_b \cdot \frac{1}{T} \sum_{t=1}^{T} \|\hat{\mathbf{T}}_{b,t} - \mathbf{T}_{b,t}^{\text{GT}}\|_1
\label{eq:traj_loss}
\end{equation}

The L1 imitation loss alone is weakly sensitive to drivable-area boundaries, especially on narrow roads, and we observe occasional waypoints slightly outside the road. We therefore add a road boundary loss as a local correction near the boundary. A point-in-polygon test first determines whether each waypoint lies inside the drivable polygon. Waypoints outside the polygon receive a constant penalty $C_{\text{out}}$, while the imitation loss supplies the corrective gradient that pulls them toward the expert trajectory. For inside waypoints, we differentiably compute $d$, the unsigned minimum distance to the nearest road boundary, and apply a piecewise penalty. Both an inverse-distance penalty and a quadratic penalty are active for $d<m_{\text{safe}}$. Only the quadratic penalty is active for $m_{\text{safe}}\le d<m_{\text{soft}}$. The penalty is zero beyond $m_{\text{soft}}$. The road term is a regularizer rather than a projection operator or safety guarantee and is defined as
\begin{equation}
\begin{aligned}
r(d) &= \frac{1}{d+\epsilon} -
        \frac{1}{m_{\text{safe}}+\epsilon},\\
q(d) &= \left(\frac{m_{\text{soft}}-d}{\Delta m}\right)^2,\\
\mathcal{L}_{\text{road}} &=
\begin{cases}
C_{\text{out}}, & \text{outside polygon},\\
r(d)+q(d), & \text{inside},\ d<m_{\text{safe}},\\
q(d), & \text{inside},\ m_{\text{safe}}\le d<m_{\text{soft}},\\
0, & \text{inside},\ d\ge m_{\text{soft}}.
\end{cases}
\end{aligned}
\label{eq:road_loss}
\end{equation}
where $\Delta m = m_{\text{soft}} - m_{\text{safe}}$ and $C_{\text{out}}$ is a fixed large penalty for waypoints located outside the drivable polygon.
The overall training objective combines the physical-space losses with the latent drift loss from Section~\ref{subsec:latent_drift} and is written as $\mathcal{L}_{\text{total}} = \lambda_1 \mathcal{L}_{\text{traj}} + \lambda_2 \mathcal{L}_{\text{drift}} + \lambda_3 \mathcal{L}_{\text{road}}$.
%-------------------------------------------------------------------------------

\section{EXPERIMENTAL RESULTS}
\label{sec:result}

\subsection{Experimental Setup}
\label{subsec:experimental_setup}

We train and evaluate DRIFT on NAVSIM~\cite{Dauner2024NEURIPS}, a data-driven planning benchmark constructed from nuPlan sensor logs and map annotations. NAVSIM evaluates a predicted ego trajectory through non-reactive pseudo-simulation, in which the ego vehicle is propagated with a vehicle model while the recorded traffic participants do not respond to its actions. We train on \textit{navtrain} and report results on \textit{navtest} under both benchmark protocols. NAVSIM v1 combines No at-fault Collision (NC), Drivable Area Compliance (DAC), Time-to-Collision (TTC), Comfort (Comf.), and Ego Progress (EP) into the PDM Score (PDMS). NAVSIM v2 adds Driving Direction Compliance (DDC), Traffic Light Compliance (TLC), Lane Keeping (LK), History Comfort (HC), and Extended Comfort (EC), producing the Extended PDM Score (EPDMS).

To construct the unconditional prior for latent-space drift (Section~\ref{subsec:latent_drift}), we extract structured road graph data for each training scene. We then test whether ground-truth trajectories from other scenes satisfy the road boundary constraints of the current scene. Trajectories that remain within the drivable area are retained, and 32 diverse trajectories are selected as the unconditional sample set for that scene. This set gives the decoder a map-compatible geometric prior, independent of the specific expert demonstration.

\begin{figure}[t]
  \centering
  \includegraphics[width=\columnwidth]{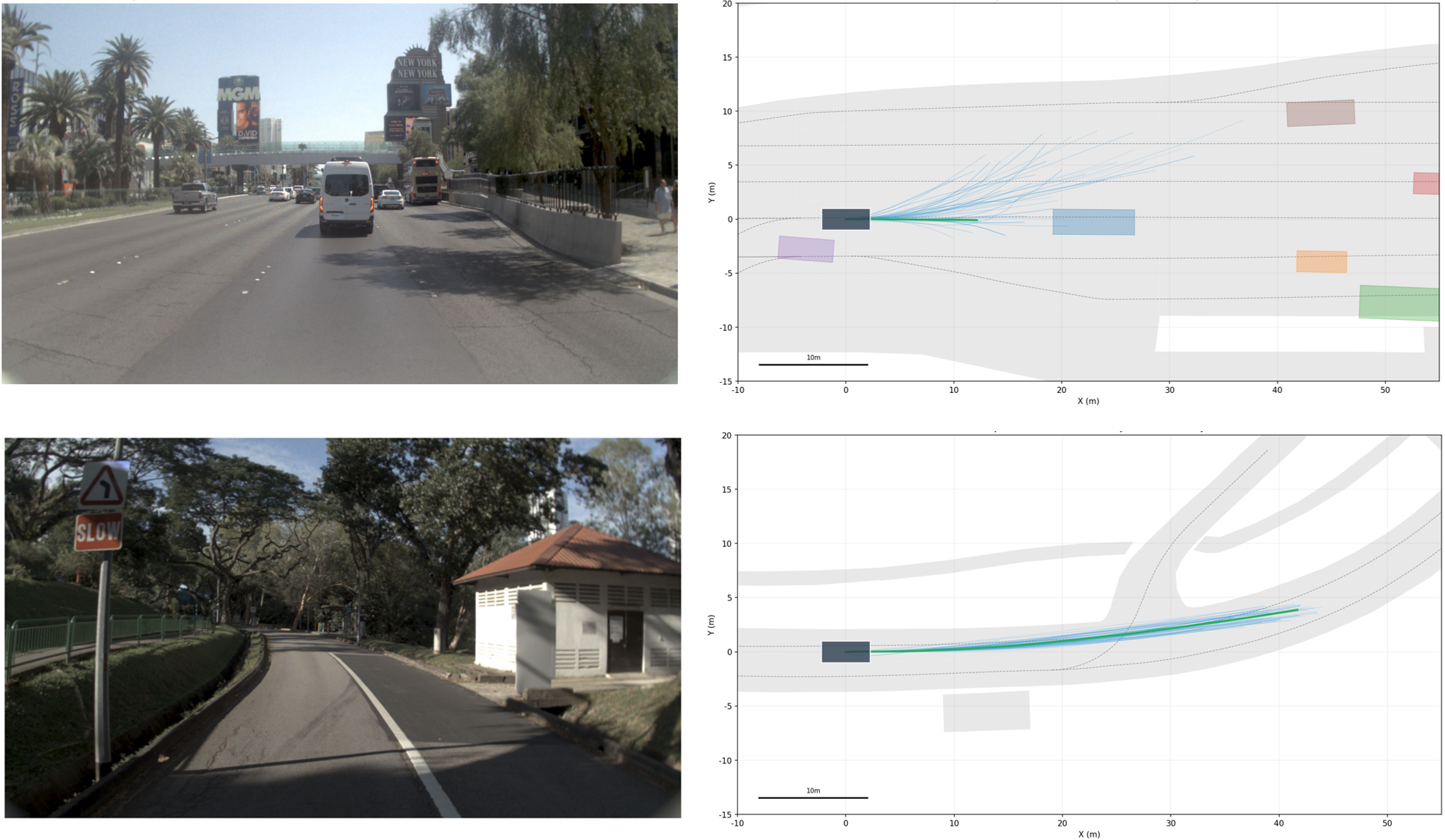}
  \caption{Unconditional trajectory samples for two representative scenes. The top row shows a multi-lane straight road where the unconditional set contains lane-keeping and lane-changing maneuvers. The bottom row shows a narrow road where the unconditional trajectories vary in speed while remaining within lane boundaries. In both cases, the sampled trajectories are map-compatible geometric priors.}
  \label{fig:unconditional}
\end{figure}

Fig.~\ref{fig:unconditional} shows two representative scenes and their unconditional trajectory sets. On the multi-lane straight road, the prior includes lane keeping and lane changes. On the narrow road, the selected trajectories mainly vary in speed and stay within the constrained lane boundaries. These examples illustrate how the unconditional samples encode local drivable-area geometry.

\subsection{Model Training}

Before training, we fit the PCA basis used in the DRIFT Decoder (Section~\ref{subsec:draft_decoder}) on the position components of ground-truth trajectories from \textit{navtrain}. Each trajectory contains eight two-dimensional waypoints, giving a 16-dimensional input vector, which is projected to an eight-dimensional latent space. The resulting PCA basis, mean, and inverse transform remain fixed throughout both training stages.

Training has two stages. In the first stage, the V-JEPA-pretrained ViT encoder is frozen, and only the DRIFT Decoder, including the PCA-coefficient MLP $f_\psi$, is trained for 50 epochs with the drift loss (Section~\ref{subsec:latent_drift}). For each scene, we generate 32 proposals from independent Gaussian noise vectors and randomly sampled drift intensities $\alpha\in[0,1]$. This stage learns scene-conditioned proposal generation without aggregation objectives. In the second stage, the ViT encoder is unfrozen and all learnable components, including the ViT, Decoder, $f_\psi$, and the newly added Aggregation Head, are jointly optimized for another 50 epochs. The objective combines the drift loss, the imitation loss $\mathcal{L}_{\text{traj}}$ (Eq.~\ref{eq:traj_loss}), and the road boundary loss $\mathcal{L}_{\text{road}}$ (Eq.~\ref{eq:road_loss}). The drift loss backpropagates through the scene-conditioned Decoder and $f_\psi$ to the ViT. The imitation and road-boundary losses update the Aggregation Head and propagate through both its scene-attention path and the hidden proposal features to the ViT and Decoder. The same 32-proposal sampling strategy is used during this stage. Both stages use AdamW with a cosine annealing schedule.

DRIFT predicts 8 waypoints over a 4 s horizon, with a temporal spacing of 0.5 s. At inference time, we replace random $\alpha$ sampling with two fixed drift intensities, $\alpha=0.5$ and $\alpha=0.9$. The decoder samples 32 Gaussian noise vectors at $\alpha=0.5$ and 16 at $\alpha=0.9$, giving 48 hidden proposal features in one batched generation pass. The Aggregation Head operates directly on the hidden proposal features rather than on their PCA-decoded trajectories. A separate PCA branch maps these features through the PCA-coefficient MLP $f_\psi$ to position trajectories, which are used to interpret and visualize the decoder hypotheses. In the first stage, the drift loss has weight 1. In the second stage, the imitation loss is the main objective with weight 1, the drift loss weight is reduced to 0.1, and the road boundary loss weight is 0.2. Thus, the boundary term acts as an auxiliary geometric regularizer rather than the primary training signal.

\begin{table*}[t]
  \caption{PDMS comparison among non-scorer-based planners on NAVSIM v1 navtest.}
  \label{tab:pdm_comparison}
  \centering
  \begin{tabular*}{\textwidth}{@{\extracolsep{\fill}} l c c c c c c @{}}
    \toprule
    \textbf{Method} & \textbf{NC}$\uparrow$ & \textbf{DAC}$\uparrow$ & \textbf{EP}$\uparrow$ & \textbf{TTC}$\uparrow$ & \textbf{Comf.}$\uparrow$ & \textbf{PDMS}$\uparrow$ \\
    \midrule
    DiffusionDrive~\cite{liao2025diffusiondrive}   & 98.2 & 96.2 & 82.2 &  94.7 & \textbf{100} & 88.1 \\
    WoTE~\cite{li2025end}   & 98.5 & 96.8 &  81.9 &  94.9 & 99.9 & 88.3 \\
    World4Drive~\cite{zheng2025world4drive}   & 97.4 & 94.3 & 79.9 &  92.8 & \textbf{100} & 85.1 \\
    TransFuser~\cite{chitta2022transfuser}   & 97.7 & 92.8 & 79.2 &  92.8 & \textbf{100} & 84.0 \\
    Epona~\cite{zhang2025epona}   & 97.9 & 95.1 & 80.4 &  93.8 & 99.9 & 86.2 \\
    MeanFuser~\cite{wang2026meanfuser}   & 98.6 & 97.0 & 82.8 & 95.0 & \textbf{100} & 89.0 \\
    Drive-JEPA~\cite{wang2026drive} & \textbf{98.7} & 96.2 & 82.9 & \textbf{95.5} & \textbf{100} & 89.0 \\
    \midrule
    \textbf{DRIFT} & 98.6 & \textbf{97.6} & \textbf{84.0} & 94.6 & \textbf{100} & \textbf{89.6} \\
    \bottomrule
  \end{tabular*}
\end{table*}

\subsection{Main Results}

\begin{table*}[t]
  \caption{EPDMS comparison among non-scorer-based planners on NAVSIM v2 navtest. $^\dagger$ denotes results reproduced with the released DiffusionDrive checkpoint.}
  \label{tab:epdms_comparison}
  \centering
  \begin{tabular*}{\textwidth}{@{\extracolsep{\fill}} l c c c c c c c c c c @{}}
    \toprule
    \textbf{Method} & \textbf{NC}$\uparrow$ & \textbf{DAC}$\uparrow$ & \textbf{DDC}$\uparrow$ & \textbf{TLC}$\uparrow$ & \textbf{EP}$\uparrow$ & \textbf{TTC}$\uparrow$ & \textbf{LK}$\uparrow$ & \textbf{HC}$\uparrow$ & \textbf{EC}$\uparrow$ & \textbf{EPDMS}$\uparrow$ \\
    \midrule
    Ego Status MLP~\cite{li2024ego}   & 93.1 & 77.9 & 92.7 & 99.6 & 86.0 & 91.5 & 89.4 & \textbf{98.3} & 85.4 & 64.0 \\
    TransFuser~\cite{chitta2022transfuser}      & 96.9 & 89.9 & 97.8 & 99.7 & 87.1 & 95.4 & 92.7 & \textbf{98.3} & 87.2 & 76.7 \\
    Hydra-MDP++~\cite{li2025hydra}      & 97.2 & 97.5 & 99.4 & 99.6 & 83.1 & 96.5 & 94.4 & 98.2 & 70.9 & 81.4 \\
    DriveSuprim~\cite{yao2026drivesuprim}      & 97.5 & 96.5 & 99.4 & 99.6 & \textbf{88.4} & 96.6 & 95.5 & \textbf{98.3} & 77.0 & 83.1 \\
    DiffusionDrive$^\dagger$~\cite{liao2025diffusiondrive} & 98.2 & 96.3 & 99.4 & 99.8 & 87.4 & 97.4 & 97.0 & \textbf{98.3} & 87.7 & 88.3 \\
    MeanFuser~\cite{wang2026meanfuser}      & 98.3 & 97.2 & \textbf{99.6} & 99.8 & 87.6 & 97.4 & \textbf{97.3} & \textbf{98.3} & \textbf{88.2} & 89.5 \\
    \midrule
    \textbf{DRIFT} & \textbf{98.7} & \textbf{98.1} & 99.5 & \textbf{99.9} & 87.3 & \textbf{98.0} & 96.8 & \textbf{98.3} & 88.0 & \textbf{90.4} \\
    \bottomrule
  \end{tabular*}
\end{table*}
Tables~\ref{tab:pdm_comparison} and~\ref{tab:epdms_comparison} compare DRIFT with recent non-scorer-based planners on the 12,146 scenarios of NAVSIM \textit{navtest}. DRIFT establishes new state-of-the-art results in this setting under both protocols, reaching 89.6 PDMS and 90.4 EPDMS. It exceeds the strongest listed alternatives by 0.6 PDMS under v1 and 0.9 EPDMS under v2. The consistent ranking across both protocols is important because v2 broadens the evaluation from the five original components to additional measures of rule compliance and temporal behavior. The advantage therefore persists under a substantially wider metric suite without using evaluator-derived trajectory-quality labels for proposal aggregation.

Under v1, DRIFT leads both DAC and EP while retaining near-best collision avoidance and the saturated top comfort score. Relative to the two tied runners-up, MeanFuser and Drive-JEPA, the improvement is concentrated in road compliance and progress rather than a uniform increase in every sub-metric. DAC and EP capture complementary failure modes because the former penalizes departures from the drivable region while the latter penalizes insufficient advancement along the route. Leading both suggests that the aggregate gain is not obtained by trading progress for geometric compliance, or vice versa. DRIFT does not lead TTC, so its highest PDMS reflects a favorable overall balance rather than dominance over every safety-related measure.

The v2 results reinforce this performance profile under a broader evaluation. DRIFT ranks first not only in EPDMS but also in NC, DAC, TLC, and TTC, while remaining within half a point of the strongest entries in DDC, LK, and EC. HC is nearly saturated across the compared methods and consequently provides little discrimination. The aggregate advantage is therefore supported by several complementary aspects of collision avoidance, road compliance, traffic-rule adherence, and temporal behavior instead of being driven by one favorable component. Although DRIFT does not maximize EP under v2, its competitive progress together with the leading compliance metrics indicates that the planner does not obtain its result simply by producing conservative trajectories.

The repeated strength in DAC across both protocols is consistent with the two map-aware elements of DRIFT. The unconditional proposal set supplies hypotheses that respect the local drivable-area geometry, while the boundary loss regularizes the final plan near road edges. At the same time, the leading v1 EP and competitive v2 EP suggest that this geometric bias does not prevent effective route progress. The main comparison alone does not isolate the contribution of either component, which is examined separately in Section~\ref{subsec:ablation}.

MeanFuser provides the closest design-level comparison because it also combines one-step proposal generation with direct reconstruction rather than evaluator-supervised trajectory scoring. DRIFT improves the aggregate result over MeanFuser under both protocols. This system-level comparison supports compact latent-space drift and feature-level aggregation as an effective alternative to generating and reconstructing proposals directly in physical trajectory space.

We also profile inference latency on an NVIDIA RTX 4090 GPU. The trajectory generation module, including the DRIFT Decoder, alpha pooling, and Aggregation Head, processes all 48 proposals in 10.82 ms, whereas the full model takes 66.43 ms. The proposal-to-plan computation therefore accounts for approximately 16\% of end-to-end latency and is not the dominant runtime cost in this implementation. Batched fixed-depth generation preserves multi-hypothesis reasoning while most computation remains in the visual backbone and the rest of the end-to-end model.

\subsection{Qualitative Results}

\begin{figure*}[t]
  \centering
  \includegraphics[width=\textwidth]{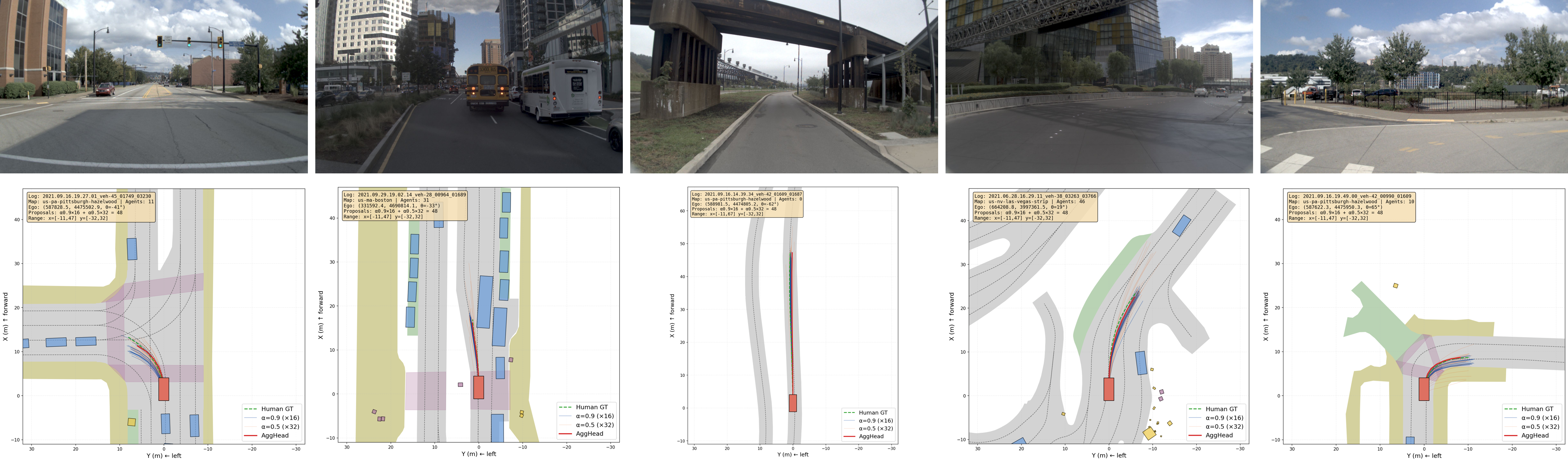}
  \caption{Planning results across five representative driving scenarios. The top row shows the front-view camera image. The bottom row shows the BEV visualization of the planned trajectory (red solid line), GT trajectory (green dashed line), and decoder proposals (orange and blue thin solid lines). From left to right, the scenarios cover a left turn, a lane change toward the left-front, straight driving on a narrow road, driving toward the right-front on a wide road, and a right turn.}
  \label{fig:planning}
\end{figure*}

Fig.~\ref{fig:planning} visualizes DRIFT on five urban maneuvers that include turning, lane changing, and straight driving under different road widths. Across these cases, the final trajectory follows the demonstrated maneuver while the intermediate proposals retain locally plausible variations for aggregation. In the left-turn and wide-road examples, proposals generated with $\alpha=0.5$ spread more than those generated with $\alpha=0.9$, which is consistent with the stronger influence of the unconditional prior at lower drift intensity. In the lane-change and narrow-road examples, both proposal groups occupy a narrower region as the scene geometry restricts the admissible motion. The examples illustrate that proposal spread varies with both $\alpha$ and scene constraints, although they do not by themselves quantify proposal diversity.

\subsection{Robustness Analysis}

The generative decoder uses Gaussian noise as input. To measure sensitivity to this noise, we evaluate the trained model on NAVSIM navtest with 10 random seeds. Table~\ref{tab:robustness} reports PDMS and EPDMS across seeds. PDMS ranges from 89.48 to 89.64, and EPDMS ranges from 90.32 to 90.38. The maximum absolute deviations from the corresponding means are 0.08 and 0.04. This experiment evaluates whether the final aggregated trajectory score remains stable under stochastic proposal sampling, ruling out favorable random seeds as the source of the observed gains.

\begin{table}[t]
  \caption{Robustness of DRIFT to inference noise across 10 random seeds.}
  \label{tab:robustness}
  \centering
  \begin{tabular*}{\columnwidth}{@{\extracolsep{\fill}} c c c @{}}
    \toprule
    \textbf{Metric} & \textbf{Seed} & \textbf{Score} \\
    \midrule
    \multirow{3}{*}{PDMS} & 16384 (Best) & 89.64 \\
                          & 456 (Worst) & 89.48 \\
                          & Avg.\ of 10 seeds & 89.56 \\
    \midrule
    \multirow{3}{*}{EPDMS} & 3333 (Best) & 90.38 \\
                           & 65535 (Worst) & 90.32 \\
                           & Avg.\ of 10 seeds & 90.36 \\
    \bottomrule
  \end{tabular*}
\end{table}

\begin{table}[t]
  \caption{Ablation of DRIFT components and proposal aggregation on NAVSIM navtest (12,146 scenarios).}
  \label{tab:component_ablation}
  \centering
  \footnotesize
  \resizebox{\columnwidth}{!}{%
  \begin{tabular}{@{} l c c c c c c @{}}
    \toprule
    \textbf{Variant} & \textbf{NC}$\uparrow$ & \textbf{DAC}$\uparrow$ & \textbf{EP}$\uparrow$ & \textbf{TTC}$\uparrow$ & \textbf{Comf.}$\uparrow$ & \textbf{PDMS}$\uparrow$ \\
    \midrule
    \textbf{Full DRIFT} & 98.6 & 97.6 & 84.0 & 94.6 & 100.0 & \textbf{89.6} \\
    Mean of 48 proposals & 96.6 & 94.7 & 81.6 & 90.8 & 99.9 & 85.1 \\
    w/o Road & 98.6 & 97.6 & 83.3 & 95.0 & 100.0 & 89.4 \\
    w/o Decoder & 98.4 & 97.6 & 83.1 & 94.3 & 100.0 & 89.2 \\
    w/o Decoder and Road & 98.6 & 97.2 & 83.8 & 94.3 & 100.0 & 89.1 \\
    \bottomrule
  \end{tabular}%
  }
\end{table}

\subsection{Ablation Study}
\label{subsec:ablation}

We conduct ablations on NAVSIM navtest to evaluate the proposal-to-plan rule and the main model components. For the aggregation baseline, we replace the learned Aggregation Head at inference time with the arithmetic mean of the same 48 PCA-decoded proposals, consisting of 32 proposals at $\alpha=0.5$ and 16 proposals at $\alpha=0.9$. The remaining configurations remove the DRIFT Decoder, the road boundary loss $\mathcal{L}_{\text{road}}$, or both components. Table~\ref{tab:component_ablation} reports the results.

Replacing learned aggregation with the mean of the same 48 proposals reduces PDMS from 89.6 to 85.1. The decrease appears across NC, DAC, EP, and TTC rather than in only one component, with the largest gaps occurring in TTC and DAC. This result indicates that generating plausible proposals is not sufficient because an unweighted physical-space mean cannot account for their scene-dependent relevance and may blend distinct motion hypotheses. The Aggregation Head instead conditions proposal features on the scene, navigation command, and ego state before directly decoding the executable trajectory.

The component removals produce smaller but consistent changes in aggregate PDMS. Adding the Decoder improves PDMS by 0.4 points when the road term is enabled and by 0.3 points without it. The road term contributes 0.2 and 0.1 points with and without the Decoder, respectively. The metric-level changes are not additive, as the largest EP gain from the Decoder appears with the road term while the DAC gain from the road term is more visible without the Decoder. We therefore treat the two components as interacting parts of the integrated planner rather than as isolated mechanisms for progress and safety.

%===============================================================================

\section{CONCLUSION}
\label{sec:conclusion}

We presented DRIFT, a fixed-depth end-to-end planner that combines one-step latent proposal generation with scene-aware aggregation. The DRIFT Decoder generates 48 intermediate proposals in a single batched pass, using 32 samples at $\alpha=0.5$ and 16 samples at $\alpha=0.9$, and the Aggregation Head fuses them into one executable trajectory without evaluator-labeled trajectory-score pairs. The final output is optimized with expert-trajectory imitation and a map-derived boundary regularizer, keeping proposal generation efficient without requiring trajectory-quality supervision.

On NAVSIM navtest, DRIFT reaches 89.6 PDMS and 90.4 EPDMS. The trajectory generation module, including proposal generation and aggregation, runs in 10.82 ms on an RTX 4090. These results support evaluator-label-free aggregation as a practical alternative for efficient proposal-based planning. Future work should evaluate the method in pseudo-closed-loop and interactive settings and should measure proposal diversity and aggregation behavior more directly.

\bibliographystyle{IEEEtran}
\bibliography{example}  % .bib

@article{wu2022trajectory,
  title={Trajectory-guided control prediction for end-to-end autonomous driving: A simple yet strong baseline},
  author={Wu, Penghao and Jia, Xiaosong and Chen, Li and Yan, Junchi and Li, Hongyang and Qiao, Yu},
  journal={Advances in Neural Information Processing Systems},
  volume={35},
  pages={6119--6132},
  year={2022}
}

@inproceedings{yin2026diffrefiner,
  title={Diffrefiner: Coarse to fine trajectory planning via diffusion refinement with semantic interaction for end to end autonomous driving},
  author={Yin, Liuhan and Ju, Runkun and Guo, Guodong and Cheng, Erkang},
  booktitle={Proceedings of the AAAI Conference on Artificial Intelligence},
  volume={40},
  number={14},
  pages={12009--12017},
  year={2026}
}

@inproceedings{Dauner2024NEURIPS,
	title = {NAVSIM: Data-Driven Non-Reactive Autonomous Vehicle Simulation and Benchmarking},
	author = {Daniel Dauner and Marcel Hallgarten and Tianyu Li and Xinshuo Weng and Zhiyu Huang and Zetong Yang and Hongyang Li and Igor Gilitschenski and Boris Ivanovic and Marco Pavone and Andreas Geiger and Kashyap Chitta},
	booktitle = {Advances in Neural Information Processing Systems (NeurIPS)},
	year = {2024},
}

@article{madjid2025trajectory,
  title={Trajectory prediction for autonomous driving: Progress, limitations, and future directions},
  author={Madjid, Nadya Abdel and Ahmad, Abdulrahman and Mebrahtu, Murad and Babaa, Yousef and Nasser, Abdelmoamen and Malik, Sumbal and Hassan, Bilal and Werghi, Naoufel and Dias, Jorge and Khonji, Majid},
  journal={Information Fusion},
  pages={103588},
  year={2025},
  publisher={Elsevier}
}

@inproceedings{dauner2023parting,
  title={Parting with misconceptions about learning-based vehicle motion planning},
  author={Dauner, Daniel and Hallgarten, Marcel and Geiger, Andreas and Chitta, Kashyap},
  booktitle={Conference on Robot Learning},
  pages={1268--1281},
  year={2023},
  organization={PMLR}
}

@article{chitta2022transfuser,
  title={Transfuser: Imitation with transformer-based sensor fusion for autonomous driving},
  author={Chitta, Kashyap and Prakash, Aditya and Jaeger, Bernhard and Yu, Zehao and Renz, Katrin and Geiger, Andreas},
  journal={IEEE transactions on pattern analysis and machine intelligence},
  volume={45},
  number={11},
  pages={12878--12895},
  year={2022},
  publisher={IEEE}
}

@article{bardes2024revisiting,
  title={Revisiting feature prediction for learning visual representations from video},
  author={Bardes, Adrien and Garrido, Quentin and Ponce, Jean and Chen, Xinlei and Rabbat, Michael and LeCun, Yann and Assran, Mahmoud and Ballas, Nicolas},
  journal={arXiv preprint arXiv:2404.08471},
  year={2024}
}

@inproceedings{jia2023driveadapter,
  title={Driveadapter: Breaking the coupling barrier of perception and planning in end-to-end autonomous driving},
  author={Jia, Xiaosong and Gao, Yulu and Chen, Li and Yan, Junchi and Liu, Patrick Langechuan and Li, Hongyang},
  booktitle={Proceedings of the IEEE/CVF International Conference on Computer Vision},
  pages={7953--7963},
  year={2023}
}

@inproceedings{jiadrivetransformer,
  title={DriveTransformer: Unified Transformer for Scalable End-to-End Autonomous Driving},
  author={Jia, Xiaosong and You, Junqi and Zhang, Zhiyuan and Yan, Junchi},
  booktitle={The Thirteenth International Conference on Learning Representations}
}

@inproceedings{huang2023gameformer,
  title={Gameformer: Game-theoretic modeling and learning of transformer-based interactive prediction and planning for autonomous driving},
  author={Huang, Zhiyu and Liu, Haochen and Lv, Chen},
  booktitle={Proceedings of the IEEE/CVF International Conference on Computer Vision},
  pages={3903--3913},
  year={2023}
}

@inproceedings{wang2025comdrive,
  title={ComDrive: Comfort-Oriented End-to-End Autonomous Driving},
  author={Wang, Junming and Zhang, Xingyu and Xing, Zebin and Gu, Songen and Guo, Xiaoyang and Hu, Yang and Song, Ziying and Zhang, Qian and Long, Xiaoxiao and Yin, Wei},
  booktitle={2025 IEEE/RSJ International Conference on Intelligent Robots and Systems (IROS)},
  pages={2682--2689},
  year={2025},
  organization={IEEE}
}

@inproceedings{hu2023planning,
  title={Planning-oriented autonomous driving},
  author={Hu, Yihan and Yang, Jiazhi and Chen, Li and Li, Keyu and Sima, Chonghao and Zhu, Xizhou and Chai, Siqi and Du, Senyao and Lin, Tianwei and Wang, Wenhai and others},
  booktitle={Proceedings of the IEEE/CVF conference on computer vision and pattern recognition},
  pages={17853--17862},
  year={2023}
}

@inproceedings{casas2024detra,
  title={Detra: A unified model for object detection and trajectory forecasting},
  author={Casas, Sergio and Agro, Ben and Mao, Jiageng and Gilles, Thomas and Cui, Alexander and Li, Thomas and Urtasun, Raquel},
  booktitle={European Conference on Computer Vision},
  pages={326--342},
  year={2024},
  organization={Springer}
}

@article{yang2024genad,
  title={GenAD: Generalized Predictive Model for Autonomous Driving},
  author={Yang, Jiazhi and Gao, Shenyuan and Qiu, Yihang and Chen, Li and Li, Tianyu and Dai, Bo and Chitta, Kashyap and Wu, Penghao and Zeng, Jia and Luo, Ping and others},
  journal={arXiv preprint arXiv:2403.09630},
  year={2024}
}

@inproceedings{liao2025diffusiondrive,
  title={Diffusiondrive: Truncated diffusion model for end-to-end autonomous driving},
  author={Liao, Bencheng and Chen, Shaoyu and Yin, Haoran and Jiang, Bo and Wang, Cheng and Yan, Sixu and Zhang, Xinbang and Li, Xiangyu and Zhang, Ying and Zhang, Qian and others},
  booktitle={Proceedings of the Computer Vision and Pattern Recognition Conference},
  pages={12037--12047},
  year={2025}
}

@inproceedings{zheng2025diffusion,
  title={Diffusion-based planning for autonomous driving with flexible guidance},
  author={Zheng, Yinan and Liang, Ruiming and Zheng, Kexin and Zheng, Jinliang and Mao, Liyuan and Li, Jianxiong and Gu, Weihao and Ai, Rui and Li, Shengbo and Zhan, Xianyuan and others},
  booktitle={International Conference on Learning Representations},
  volume={2025},
  pages={37207--37227},
  year={2025}
}

@article{li2025generalized,
  title={Generalized trajectory scoring for end-to-end multimodal planning},
  author={Li, Zhenxin and Yao, Wenhao and Wang, Zi and Sun, Xinglong and Chen, Joshua and Chang, Nadine and Shen, Maying and Wu, Zuxuan and Lan, Shiyi and Alvarez, Jose M},
  journal={arXiv preprint arXiv:2506.06664},
  year={2025}
}

@article{chen2025hipro,
  title={HiPro-AD: Sparse Trajectory Transformer for End-to-End Autonomous Driving with Hybrid Spatiotemporal Attention},
  author={Chen, Bing and Wang, Gaopeng and Yang, Jiandong and Huang, Shaoliang and Qian, Xinhe and Huang, Bin and Guo, Guanlun},
  journal={Sensors (Basel, Switzerland)},
  volume={26},
  number={1},
  pages={185},
  year={2025}
}

@article{chen2024end,
  title={End-to-end autonomous driving: Challenges and frontiers},
  author={Chen, Li and Wu, Penghao and Chitta, Kashyap and Jaeger, Bernhard and Geiger, Andreas and Li, Hongyang},
  journal={IEEE Transactions on Pattern Analysis and Machine Intelligence},
  volume={46},
  number={12},
  pages={10164--10183},
  year={2024},
  publisher={IEEE}
}

@article{wang2026drive,
  title={Drive-JEPA: Video JEPA Meets Multimodal Trajectory Distillation for End-to-End Driving},
  author={Wang, Linhan and Yang, Zichong and Bai, Chen and Zhang, Guoxiang and Liu, Xiaotong and Zheng, Xiaoyin and Long, Xiao-Xiao and Lu, Chang-Tien and Lu, Cheng},
  journal={arXiv preprint arXiv:2601.22032},
  year={2026}
}

@article{deng2026generative,
  title={Generative Modeling via Drifting},
  author={Deng, Mingyang and Li, He and Li, Tianhong and Du, Yilun and He, Kaiming},
  journal={arXiv preprint arXiv:2602.04770},
  year={2026}
}

@inproceedings{li2025end,
  title={End-to-end driving with online trajectory evaluation via bev world model},
  author={Li, Yingyan and Wang, Yuqi and Liu, Yang and He, Jiawei and Fan, Lue and Zhang, Zhaoxiang},
  booktitle={Proceedings of the IEEE/CVF International Conference on Computer Vision},
  pages={27137--27146},
  year={2025}
}

@inproceedings{zheng2025world4drive,
  title={World4drive: End-to-end autonomous driving via intention-aware physical latent world model},
  author={Zheng, Yupeng and Yang, Pengxuan and Xing, Zebin and Zhang, Qichao and Zheng, Yuhang and Gao, Yinfeng and Li, Pengfei and Zhang, Teng and Xia, Zhongpu and Jia, Peng and others},
  booktitle={Proceedings of the IEEE/CVF International Conference on Computer Vision},
  pages={28632--28642},
  year={2025}
}

@inproceedings{zhang2025epona,
  title={Epona: Autoregressive diffusion world model for autonomous driving},
  author={Zhang, Kaiwen and Tang, Zhenyu and Hu, Xiaotao and Pan, Xingang and Guo, Xiaoyang and Liu, Yuan and Huang, Jingwei and Yuan, Li and Zhang, Qian and Long, Xiao-Xiao and others},
  booktitle={Proceedings of the IEEE/CVF International Conference on Computer Vision},
  pages={27220--27230},
  year={2025}
}

@inproceedings{wang2026meanfuser,
  title={Meanfuser: Fast one-step multi-modal trajectory generation and adaptive reconstruction via meanflow for end-to-end autonomous driving},
  author={Wang, Junli and Zheng, Yinan and Liu, Xueyi and Xing, Zebin and Li, Pengfei and Ma, Kun and Ye, Hangjun and Chen, Guang and Li, Guang and Chen, Long and others},
  booktitle={Proceedings of the IEEE/CVF Conference on Computer Vision and Pattern Recognition},
  pages={17884--17893},
  year={2026}
}

@inproceedings{li2024ego,
  title={Is ego status all you need for open-loop end-to-end autonomous driving?},
  author={Li, Zhiqi and Yu, Zhiding and Lan, Shiyi and Li, Jiahan and Kautz, Jan and Lu, Tong and Alvarez, Jose M},
  booktitle={Proceedings of the IEEE/CVF Conference on Computer Vision and Pattern Recognition},
  pages={14864--14873},
  year={2024}
}

@article{li2025hydra,
  title={Hydra-mdp++: Advancing end-to-end driving via expert-guided hydra-distillation},
  author={Li, Kailin and Li, Zhenxin and Lan, Shiyi and Xie, Yuan and Zhang, Zhizhong and Liu, Jiayi and Wu, Zuxuan and Yu, Zhiding and Alvarez, Jose M},
  journal={arXiv preprint arXiv:2503.12820},
  year={2025}
}

@inproceedings{yao2026drivesuprim,
  title={Drivesuprim: Towards precise trajectory selection for end-to-end planning},
  author={Yao, Wenhao and Li, Zhenxin and Lan, Shiyi and Wang, Zi and Sun, Xinglong and Alvarez, Jose M and Wu, Zuxuan},
  booktitle={Proceedings of the AAAI Conference on Artificial Intelligence},
  volume={40},
  number={14},
  pages={11910--11918},
  year={2026}
}

@inproceedings{ke2026grade,
  title={GRADE: Guiding Realistic Autonomous Driving with Adaptive Trajectory Evolution},
  author={Ke, Zehong and Liu, Zhiyuan and Wang, Yuning and Li, Jinhao and Jiang, Junkai and Jiang, Yanbo and Xu, Zhenhua and Wang, Jianqiang},
  booktitle={Proceedings of the IEEE/CVF Conference on Computer Vision and Pattern Recognition},
  pages={1029--1038},
  year={2026}
}

\end{document}